\documentclass[10pt,conference]{IEEEtran}

\usepackage{comment}
\usepackage[shortlabels]{enumitem}

\usepackage{xcolor}
\usepackage{cite}
\usepackage{hyperref}
\ifCLASSINFOpdf
   \usepackage[pdftex]{graphicx}
   \graphicspath{{figs/}}
   \DeclareGraphicsExtensions{.pdf,.jpeg,.png}
\else
   \usepackage[dvips]{graphicx}
   \graphicspath{{../figs/}}
   \DeclareGraphicsExtensions{.eps}
\fi
\usepackage[cmex10]{amsmath}
\usepackage{amsthm}

\usepackage{algorithm}
\usepackage{algpseudocode}

\usepackage{array}

\ifCLASSOPTIONcompsoc
  \usepackage[caption=false,font=normalsize,labelfont=sf,textfont=sf]{subfig}
\else
  \usepackage[caption=false,font=footnotesize]{subfig}
\fi
\usepackage{url}

\usepackage{multirow}
\usepackage{xcolor,colortbl}
\usepackage{graphicx}
\usepackage{subcaption}
\usepackage{balance}
\newif\iffinal
\finaltrue
\newcommand{\cmtid}{116}

\iffinal
\else
\usepackage[switch]{lineno}
\fi

\begin{document}
%
\title{Sampling Strategies based on Wisdom of Crowds for Amazon Deforestation Detection}


\iffinal

\author{\IEEEauthorblockN{Hugo Resende, Eduardo B. Neto, F\'abio A. M. Cappabianco, \'Alvaro L. Fazenda, Fabio A. Faria}
\IEEEauthorblockA{Institute of Science and Technology -- Universidade Federal de S\~ao Paulo, S\~ao Jos\'e dos Campos, SP, Brazil \\
E-mail: \{hresende, ebneto, cappabianco, alvaro.fazenda, ffaria\}@unifesp.br}
}


%
\else
  \author{SIBGRAPI Paper ID: \cmtid \\ }
  \linenumbers
\fi

\maketitle

\begin{abstract}
Conserving tropical forests is highly relevant socially and ecologically because of their critical role in the global ecosystem. However, the ongoing deforestation and degradation affect millions of hectares each year, necessitating government or private initiatives to ensure effective forest monitoring. In April 2019, a project based on Citizen Science and Machine Learning models called ForestEyes (FE) was launched with the aim of providing supplementary data to assist experts from government and non-profit organizations in their deforestation monitoring efforts. Recent research has shown that labeling FE project volunteers/citizen scientists helps tailor machine learning models. In this sense, we adopt the FE project to create different sampling strategies based on the wisdom of crowds to select the most suitable samples from the training set to learn an SVM technique and obtain better classification results in deforestation detection tasks. In our experiments, we can show that our strategy based on user entropy-increasing achieved the best classification results in the deforestation detection task when compared with the random sampling strategies, as well as, reducing the convergence time of the SVM technique.



\end{abstract}

\IEEEpeerreviewmaketitle

\section{Introduction}

Tropical forests are biomes that perform indispensable activities for maintaining life and the planet's health. One of the most important biomes is the Brazilian Legal Amazon, which, due to its vast extent, harbors an unparalleled diversity of life, from plant and animal species to indigenous peoples whose cultures are intrinsically linked to it. In addition to its biological and cultural importance, the Amazon exerts significant influence on global climatic systems, regulating precipitation patterns and stabilizing the climate. However, rampant deforestation, driven by agricultural expansion, mining, and illegal activities, severely threatens the integrity of this vital ecosystem. Recent studies indicate that large areas of the Amazon have already been deforested, resulting in irreparable biodiversity losses and contributing to high carbon emissions and global climate change \cite{lapola2023drivers, reis2023econometric}. In this context, it is necessary to propose initiatives to protect and preserve the Amazon so that present and future generations do not suffer these negative consequences.

To combat deforestation and degradation of tropical forests, several monitoring programs have been developed and implemented \cite{pro2023global, diniz2015deter}. In the case of the Brazilian Legal Amazon, the Satellite Monitoring Program for the Brazilian Amazon (PRODES), coordinated by the National Institute for Space Research (INPE) of Brazil, stands out~\cite{inpe-prodes}. PRODES utilizes satellite data to generate precise annual estimates of deforestation rates, allowing for the detailed identification and mapping of affected areas. The importance of this program lies in the accuracy and reliability of the data provided, which are crucial for the formulation of effective public policies, environmental enforcement, and the implementation of conservation actions. Additionally, by providing updated information on deforestation, PRODES significantly contributes to transparency and public awareness, playing an essential role in the preservation of the tropical forests of the Legal Amazon. Other complementary initiatives include reforestation and restoration projects, environmental education campaigns through, for example, citizen science projects, and the creation of conservation units, all aimed at protecting this vital ecosystem \cite{da2021overview}.

An innovative project proposed in recent years is the ForestEyes \cite{dallaqua2019foresteyes, dallaqua2021foresteyes}. In this project, campaigns are designed and launched focusing on regions with potential for deforestation in the Brazilian Legal Amazon, where non-specialist volunteers actively participate in labeling image samples in the famous Zooniverse platform. Each small area (image segment representing a task) of remote sensing images, after thorough analysis, is manually classified by volunteers as forest or deforestation. At the end of each campaign (a set of tasks), these areas constitute a powerful labeled dataset that can be used, for example, to train machine learning (ML) models aiming to automatically identify possible recent deforestation hotspots in regions of the Brazilian Legal Amazon. Although datasets labeled by volunteers constitute valuable material, selecting quality samples for effective training of an ML model, identifying noisy samples, and creating robust training set~\cite{dallaqua_grsl2020}, the procedure is quite challenging, as the labels may not precisely represent the majority classes of the segments. With this in mind, measures of response variability, such as Shannon's entropy, can be used to assess how reliable the labels assigned by volunteers are to the segments. Furthermore, by calculating the entropy for each segment, it becomes feasible to determine the degree of classification difficulty for each task, or the confidence in labeling, based on human cognition. For example, segments with higher entropy indicate a higher level of classification difficulty. Given that there are progressive machine learning strategies, it is possible to associate segments with their respective entropies to sample groups that will be used incrementally in model training. This approach allows models to adapt gradually, starting with simpler examples and advancing to more complex ones (or vice versa), thereby optimizing the learning process and improving the accuracy and robustness of the final model~\cite{wang2021survey}.

Given that the ForestEyes citizen science campaigns benefit from the massive participation of volunteers, a fundamental characteristic of the Wisdom of the Crowds principle \cite{CitizenScienceandCrowdScience}, this study aims to investigate the effectiveness of adopting user-based entropy as measures of classification difficulty, combined with a machine learning model. The goal is to assess whether this approach is efficient in detecting deforestation in the Brazilian Legal Amazon. 


\section{Background}

In this section, we will present a theoretical foundation for the technologies, techniques, and metrics used in this research. Specifically, we will discuss the main characteristics of the Sentinel-2 satellite as well as address the PRODES project. Additionally, we will elaborate on the SLIC and MaskSLIC segmentation algorithms, and we will introduce a segment selection metric known as Homogeneity Rate and Shannon’s entropy, which measures uncertainty applied in data analysis. Finally, the Haralick texture features will be described, 
as well as the SVM classifier and its approach to building solutions.


\subsection{Sentinel-2 Satellite}


Sentinel-2 satellite is part of a program known as Copernicus, managed by the European Space Agency (ESA), and comprises two satellites, Sentinel-2A and Sentinel-2B, launched in 2015 and 2017, respectively. These satellites are equipped with the MultiSpectral Instrument (MSI), which captures images in 13 spectral bands, ranging from the visible spectrum to shortwave infrared. The spatial resolution of Sentinel-2 is 10 meters in the visible and near-infrared bands and 20 meters in the shortwave infrared bands. This enables detailed analysis for a variety of applications, such as agriculture, natural disaster management, and water resource monitoring. Furthermore, Sentinel-2 has a revisit frequency of 5 days, providing short-term observation essential for environmental monitoring and scientific research~\cite{drusch2012sentinel,main2017sen2cor}. 

\subsection{PRODES Project}

The PRODES Project, developed by the National Institute for Space Research (INPE) of Brazil \cite{inpe-prodes}, has been monitoring deforestation in the Legal Amazon since 1988. It measures annual deforestation rates for the period between August of one year and July of the following year. Initially intended for governmental use, PRODES evolved into a platform accessible to the public, especially after the introduction of digital PRODES in 2003. This update included the dissemination of digital maps, making deforestation data more comprehensible for all stakeholders \cite{valeriano2004monitoring}.

\textcolor{black}{Unlike other Brazilian monitoring systems, such as DETER-B, PRODES uses high spatial resolution data captured by sensors like Sentinel-2, with some bands reaching 10 meters. Additionally, its data is considered more reliable, as it is meticulously analyzed throughout the year by experts in forest remote sensing.} A key feature of PRODES is its database of multi-temporal satellite images, available to the public, allowing for a comprehensive historical analysis of deforestation in the Amazon. It uses data from satellites such as Landsat-8, Sentinel-2 \cite{drusch2012sentinel}, and CBERS-4 \cite{epiphanio2011cbers} to provide high-resolution images of the Legal Amazon. These images serve as a reference for evaluating the quality of segmentation in deforestation studies.

\begin{figure}[ht!]
\centering
\includegraphics[width=0.4\textwidth]{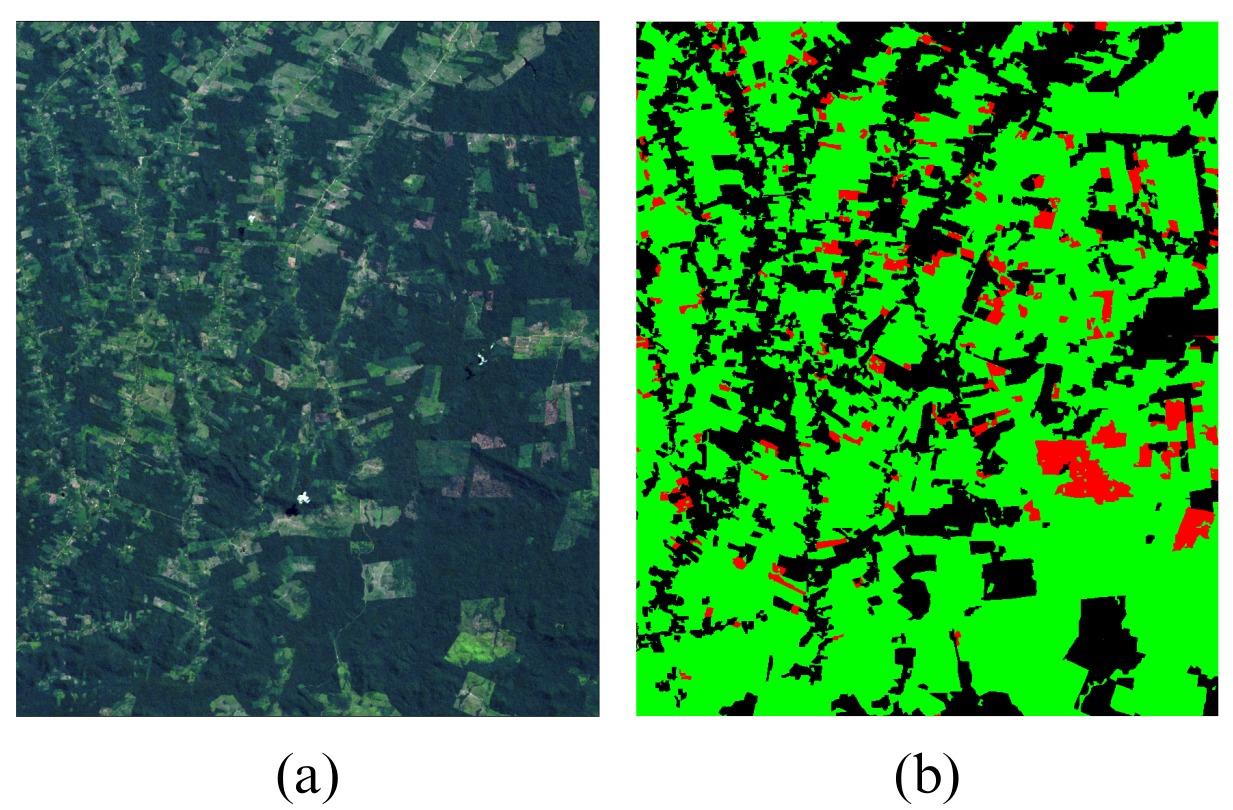}
\caption{Study Area and its respective ground truth by PRODES.}
\label{fig:sa-x08}
\end{figure}

To illustrate a geographic forest area with small deforested regions identified by PRODES, Figure \ref{fig:sa-x08} presents two images. In the first image (Figure \ref{fig:sa-x08}(a)), a study area used in the present work is shown, while the second image (Figure \ref{fig:sa-x08}(b)) displays the corresponding actual deforested area, known as PRODES ground truth. In this representation, red pixels indicate recently deforested areas (non-forest), green pixels correspond to forest regions, and black pixels represent areas that should not be analyzed during segmentation, such as hydrography and consolidated deforestation.
\vspace{-0.25cm}

\vspace{-0.5cm}
\subsection{SLIC Algorithms}
The Simple Linear Iterative Clustering (SLIC) algorithm is a method for segmenting color images represented in the CIELAB color space, based on generating superpixels using the K-means clustering algorithm. Initially, it takes parameters such as the desired number of superpixels ($k$) and a compactness control ($m$) to ensure uniformity in the size and shape of the superpixels. Its efficiency is notable because its computational complexity is linear in relation to the number of pixels ($N$), as the search is limited to the size of the superpixels \cite{achanta2012slic}.

The segmentation process begins by converting the image to the CIELAB color space and creating $k$ clusters of size $\sqrt{\frac{N}{k}}$ pixels. Each superpixel is represented by a centroid that contains the average values of the color components and spatial coordinates. In each iteration, every pixel is associated with the nearest centroid within a delimited region, and the centroids are updated with the average of the pixels belonging to the superpixel, repeating the process for a fixed number of iterations until stability is achieved.

On the other hand, MaskSLIC is an extension of SLIC developed to handle irregular masks, or regions of interest (RoIs), in the image \cite{irving2016maskslic}. Unlike SLIC, which uniformly distributes seed points across the image, MaskSLIC positions these points only within the RoI using an Euclidean distance transform to ensure effective coverage. After defining the seed points, MaskSLIC applies the SLIC procedure within the mask, ensuring that superpixels are evenly distributed and respect the boundaries of the RoI, providing more accurate and consistent segmentation in specific areas of interest.
\subsection{Homogeneity Rate ($HoR$)}

Currently, in the ForestEyes project, to measure the quality of segments that can be used in citizen science campaigns, a metric known as the Homogeneity Rate ($HoR$) is adopted \cite{dallaqua2021foresteyes}. Originally, this measure indicates the percentage of pixels belonging to a specific class in a binary classification. In the context of the ForestEyes project, it quantifies the percentage of pixels of the forest or non-forest classes within the segments, according to the majority class of the pixels in each segment. The value of $HoR$ for each segment can be evaluated using Equation \ref{eq:hor}.

\begin{equation}
\label{eq:hor}
HoR = \frac{max(NFP, NNP)}{NP},
\end{equation}

where, $NFP$ represent the number of forest class pixels, $NNP$ is the number of pixels in the non-forest class, and finally, $NP$ concerns the total number of pixels in the segment.

\subsection{Shannon's Entropy}

After the conclusion of each campaign, as mentioned in the introduction of this paper, the majority response from the volunteers for each segment should be calculated, given a specific number of evaluations each segment can receive. By analyzing the responses per segment, it is possible to determine the entropy for each one. In this particular study, Shannon entropy was used, which can be represented by Equation \ref{eq:shannon}.

\begin{equation}
\mathcal{H}(x) = - \sum_{i=1}^{n} p(x_i) \cdot \log_2(p(x_i)),
\label{eq:shannon}
\end{equation}

where $p(x_i)$ is the probability of the event $x_i$ occurring and $n$ is the number of classes. In this work, this means the probability of one of the classes (forest or non-forest) being chosen by the volunteer, which can be calculated by dividing the number of responses for a class $i$ by the total number of responses received for that segment. Specifically, in the present study, entropy proves to be a good measure for assessing the classification difficulty of the segments, as segments with higher entropy values tend to be more difficult for volunteers to classify (greater variability in responses).

\subsection{Haralick Texture Features}

Haralick texture descriptors pertain to a set of $14$ features that capture an image's texture based on the gray-level co-occurrence matrix (GLCM). The GLCM counts the frequency of pixel pairs with specific values, separated by a set distance and direction. Considering four directions ($0$, $45$, $90$, and $135$ degrees), four matrices are computed and normalized to form a probability matrix. From this matrix, Haralick's $14$ descriptors can be computed.

Among these descriptors, (i) \textit{Angular Second Moment} measures GLCM uniformity, indicating homogeneous textures; (ii) \textit{Contrast} evaluates intensity variation between a pixel and its neighbors, highlighting texture contrast; (iii) \textit{Correlation} measures the linear relationship between pixels, indicating dependency between pixel intensities; (iv) \textit{Sum of Squares (Variance)} represents the dispersion of values around the mean, associated with intensity diversity; (v) \textit{Inverse Difference Moment} measures texture uniformity, assigning higher values to elements near the GLCM diagonal; (vi) \textit{Sum Average} calculates the mean of summed pixel intensities, reflecting central tendency; (vii) \textit{Sum Variance} evaluates the variation of summed intensities, indicating pattern diversity; (viii) \textit{Sum Entropy} measures disorder or randomness in summed pixel intensities; (ix) \textit{Entropy} quantifies the randomness or complexity of the image texture; (x) \textit{Difference Variance} measures the dispersion of intensity differences in the GLCM; (xi) \textit{Difference Entropy} assesses disorder or randomness in intensity differences; (xii and xiii) \textit{two measures of correlation information} quantify mutual dependence between gray levels, indicating pixel intensity relationships; and (xiv) \textit{Maximal Correlation Coefficient} measures the highest correlation between pixel intensities, reflecting similarity across image regions \cite{haralick1973textural, humeau2019texture}.

\subsection{Support Vector Machine}

The Support Vector Machine (SVM) is a supervised learning algorithm used for classification and regression tasks. Its primary goal is to find an optimal hyperplane that best separates the classes in the input data. In other words, the SVM seeks to identify the decision hyperplane that maximizes the margin between the data points of different classes. This is achieved by using support vectors, which are the points closest to the decision hyperplane \cite{cortes1995support, vapnik1998statistical}.

One of the most useful features of the SVM is the so-called kernel trick, which allows the separation of classes in high-dimensional spaces. The kernel trick maps low-dimensional input data to a high-dimensional feature space without the need to explicitly compute these new dimensions, using only kernel functions such as linear and polynomial, for example. During the prediction phase, new data points are mapped to the high-dimensional space and classified based on which side of the hyperplane they lie on. Among its main advantages are the ability to efficiently handle high-dimensional data and robustness against overfitting in high-dimensional spaces due to the maximization of the margin.

\section{ForestEyes Project}
\label{sec:FE}
\begin{figure*}[!htb]
    \centering
    \includegraphics[width=0.85 \textwidth]{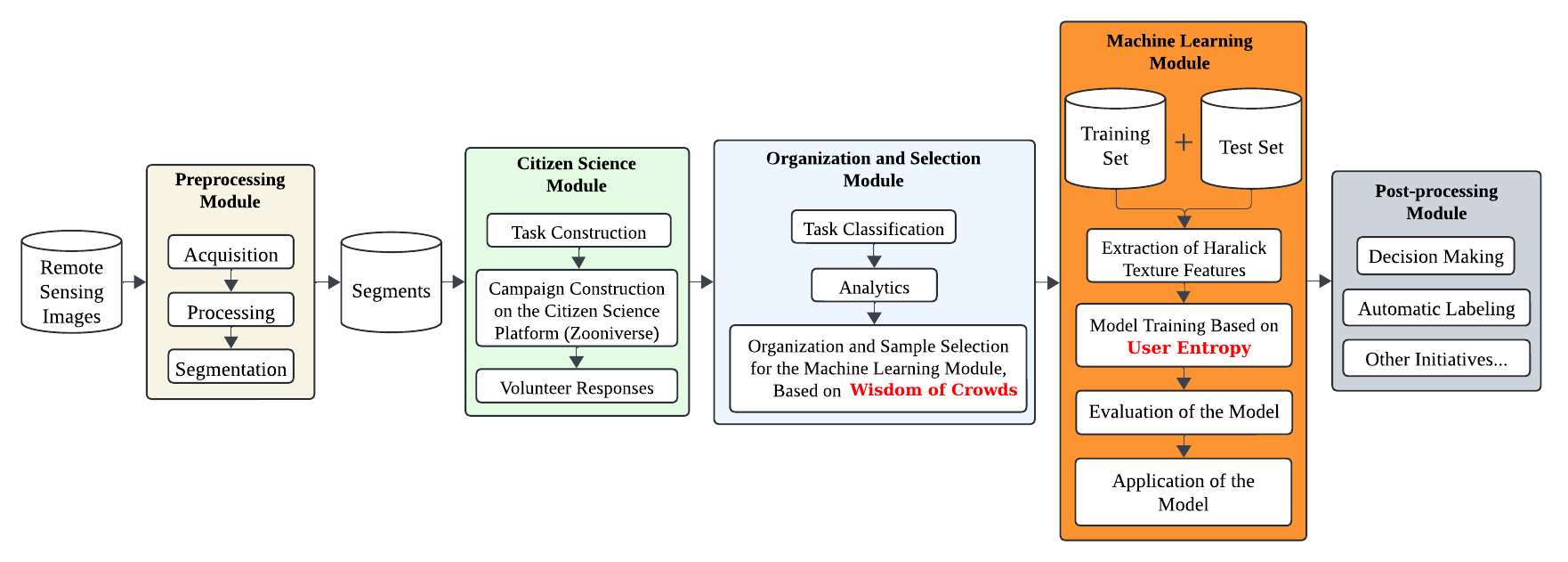}
    \caption{ForestEyes project's schematic representation. The light blue and orange modules correspond, respectively, to the modules implemented in this work.}
    \label{fig:esquemaFE}
\end{figure*}

In this section, the modules that constitute the ForestEyes project (FE) are explained in detail. To better guide the understanding of each module, Figure~\ref{fig:esquemaFE} illustrates a schematic representation of this project.

\begin{enumerate}[(a)]

\item In the first module, called the \textbf{Preprocessing Module}, the acquisition, processing, and segmentation of images from various satellites are performed. Particularly, in this research, after selecting the geographic area of interest using the engine of Sentinel-2, bands B1 (coastal aerosol), B2 (blue), B3 (green), B4 (red), B8 (vnir), B11 (swir), and B12 (swir) are collected.  
Regarding the geographic region investigated in this work, specifically, $9$ areas of interest (study areas) located geographically in a region of Brazil known as the Xingu River Basin, in the state of Pará, were collected. Together, these $9$ areas cover a total of $8,514$ hectares of land. After collecting the bands, a composition (combination) of three out of the seven bands is done before segmenting the collected area. Following some preliminary experiments with the SLIC ~\cite{achanta2012slic} and MaskSLIC~\cite{irving2016maskslic} segmenters, it was identified that using the composition of bands B4, B3, and B2 (B4B3B2), corresponding to RGB bands, generated better quality segmentations. Thus, after performing the mentioned composition, the MaskSLIC algorithm was applied so that non-interest regions, such as consolidated deforestation, were ignored during the segmentation;

\item In the \textbf{Citizen Science Module}, each segment, which can be understood in the form of a task, was visually represented in four different false-color compositions. In this sense, the compositions b8b11b4, b4b11b12, and b11b8b4 were used, along with the NDVI. In the construction of the campaign, $90$ segments with $HoR = 1$ and $90$ segments with $HoR$ between $0.7$ and $0.8$ were used, aiming to analyze the volunteers' behavior when classifying more and less homogeneous segments. After creating the tasks, the campaign (workflow) were constructed on the Zooniverse platform~\cite{simpson2014zooniverse, smith2013introduction}. During a period of approximately two weeks, volunteers' responses were collected. This methodology allowed for a detailed analysis of volunteers' performance concerning the homogeneity of the segments, providing a clearer understanding of the effectiveness of the false-color compositions and NDVI in facilitating the classification task;

\item Once the campaign is finished, the volunteers' responses are processed and analyzed in the \textbf{Organization and Selection Module}. This module assesses factors like response times and entropies to evaluate quality. It determines the specific number of responses per task. By calculating the majority response for each task, variability is measured using entropy ($\mathcal{H}$). For this project, \textcolor{black}{it was computed} an entropy value for each task to create a ranking from highest to lowest, which assesses the complexity of classification for each task. Based on this ranking, the segments constituting each task were turned into a training set for the Machine Learning Module. This approach allowed for a thorough analysis of the quality of the volunteers' classifications and helped in choosing the most suitable samples for training machine learning models;

\item In the penultimate module, the \textbf{Machine Learning Module}, training and test sets are constructed, and Haralick texture features are extracted. 
The SVM technique using linear kernel is trained in different approaches and strategies, and subsequently evaluated and applied to other regions of interest. For the training set, this study used only segments classified by volunteers as 'forest' or 'non-forest', given that the campaigns allowed for responses such as 'I cannot identify the class' and the possibility of ties in the majority response. Quantitatively, the training set comprised $96$ samples of the 'forest' class and $83$ samples of the 'non-forest' class for the Sentinel-2 campaign. For the test set, $54771$ segments of the 'forest' class and $7872$ segments of the 'non-forest' class were used, resulting in $62,643$. From these segments, $13$ Haralick texture features in four different directions were extracted, resulting in $72$ features, which served as input for the machine learning model. Regarding the methodology implemented in the training process, twenty different training sets were constructed. The first set, balanced by class, contained $5$\% of the samples with the highest entropy. Starting from the second set, an additional $5$\% of samples were included, in descending order of entropy, incrementally until the twentieth set, which contained all the samples ($100$\%). Once the training sets were constructed, three sampling strategies were implemented during the model training process (user-based entropy). These strategies are referred to in this paper as \textit{increasing}, \textit{decreasing}, and \textit{edges}. In the \textit{increasing} strategy, the training set selected of the model consisted of samples in ascending order of entropy. Conversely, in the \textit{decreasing} strategy, the sets with samples of higher entropy were selected. The \textit{edges} strategy involved selecting 2.5\% of the samples with the highest entropy and 2.5\% of the samples with the lowest entropy. 
Finally, at the end of the training, using balanced accuracy, the model was evaluated and made available for application in the automatic detection of deforestation areas;

\item While not the main focus of this study, the \textbf{Post-processing Module} includes a detailed analysis of the obtained results. These findings have the potential to serve as a crucial basis for various future applications, such as strategic decision-making and the development of effective alert systems. Additionally, in this module, based on labels assigned by volunteers in citizen science campaigns, it is anticipated that other approaches, such as automatic labeling, could be implemented in the near future.

\end{enumerate}

\section{Results and Discussion}

In this section, the results obtained in the present research are presented and discussed. 

\subsection{Sentinel-2 Campaign Analysis}

\begin{figure}[ht!]
\centering
\includegraphics[width=0.41\textwidth]{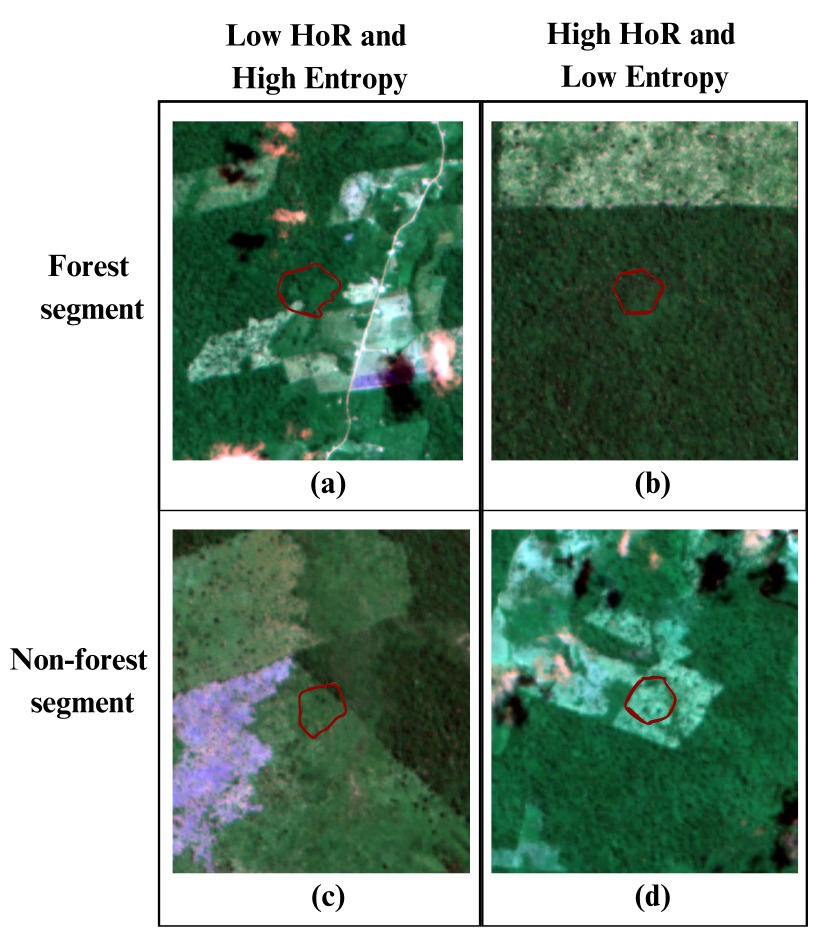}
\caption{Examples of segments (in red color) with different $HoR$ and entropy values, showing that the higher the $HoR$ value, the lower the entropy tends to be and vice versa.}
\label{fig:segments_hor_entropy}
\end{figure}

In this experiment, \textcolor{black}{it was analyzed} the behavior of user ratings through entropy and its relationship with the homogeneity coefficient of the task segment. In this sense, the Figure \ref{fig:segments_hor_entropy} illustrates segments with low and high entropy. A low entropy value generally corresponds to a high $HoR$, indicating well-defined segments, such as in Figures \ref{fig:segments_hor_entropy}b and \ref{fig:segments_hor_entropy}d, representing forest and non-forest areas, respectively. On the other hand, high entropy and low $HoR$ indicate that the image segments can contain a mix of pixels for the two classes, as shown in Figures \ref{fig:segments_hor_entropy}a and \ref{fig:segments_hor_entropy}c, leading to a poorly defined delimiting polygon. Segments with combinations of high entropy and high $HoR$, and vice versa, are harder to find in \textcolor{black}{in this research}.

\begin{figure}[ht!]
\centering
\includegraphics[width=0.49\textwidth]{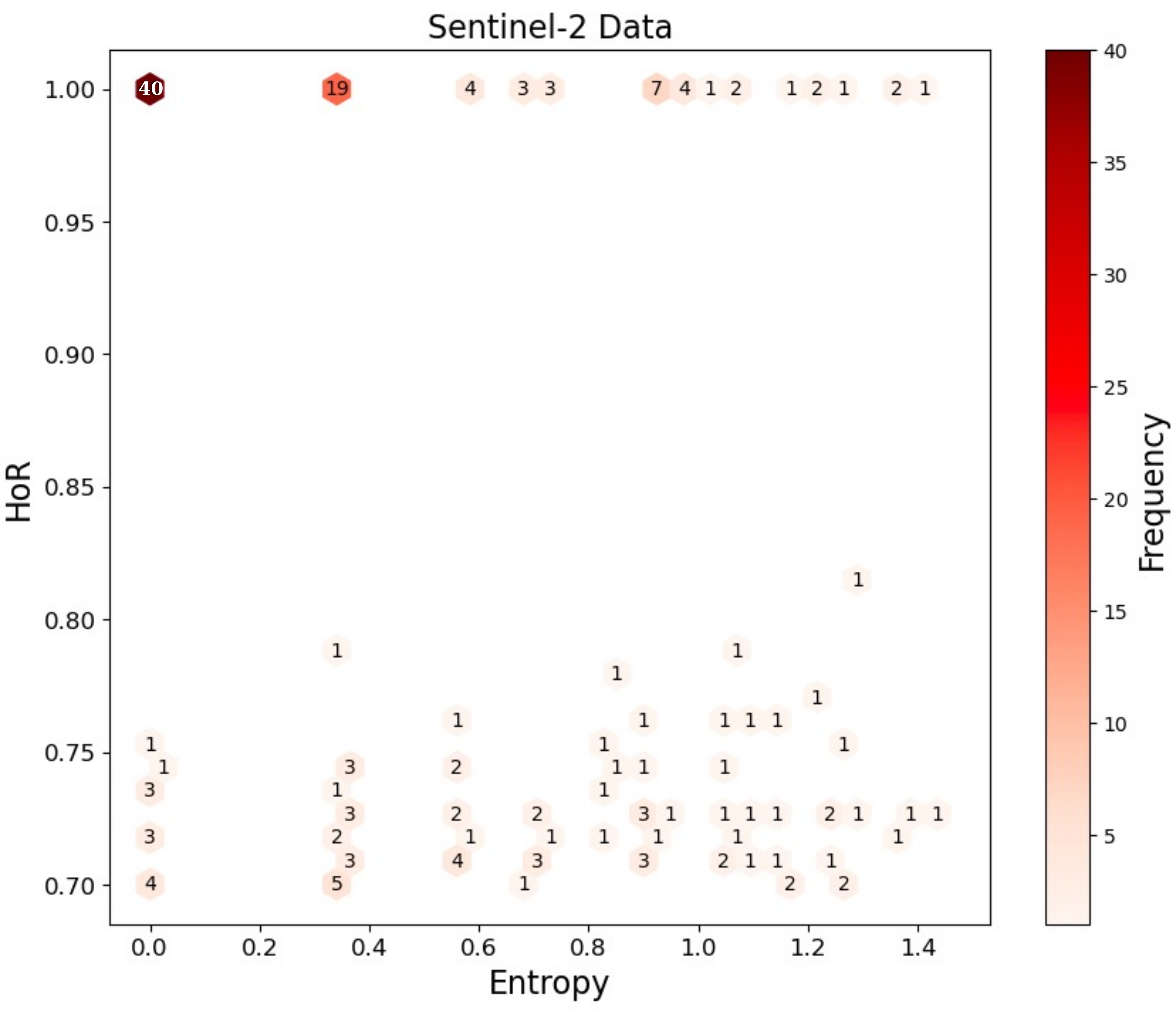}
\caption{$HoR$ $\times$ Entropy values for Sentinel-2 Campaign.}
\label{fig:hor_entropy_sentinel}
\end{figure}

According to this behavior, in Figure \ref{fig:hor_entropy_sentinel}, it is possible to analyze the relationship between Entropy and $HoR$ for all segments used in the training data set. Most of the perfect segments ($HoR = 1.0$), \textcolor{black}{presented} in the superior line, could be considered as high confidence ($\mathcal{H} = 0.0$) by the volunteers. Otherwise, segments with \textcolor{black}{lower} $HoR$, populating the inferior lines in the same figure, are \textcolor{black}{presented} close to the right inferior corner, meaning a low confident volunteers' classification ($\mathcal{H} > 0.8$).

\subsection{Comparative Analysis among Sampling Strategies}

\begin{figure}[ht!]
\centering
\includegraphics[width=0.49\textwidth]{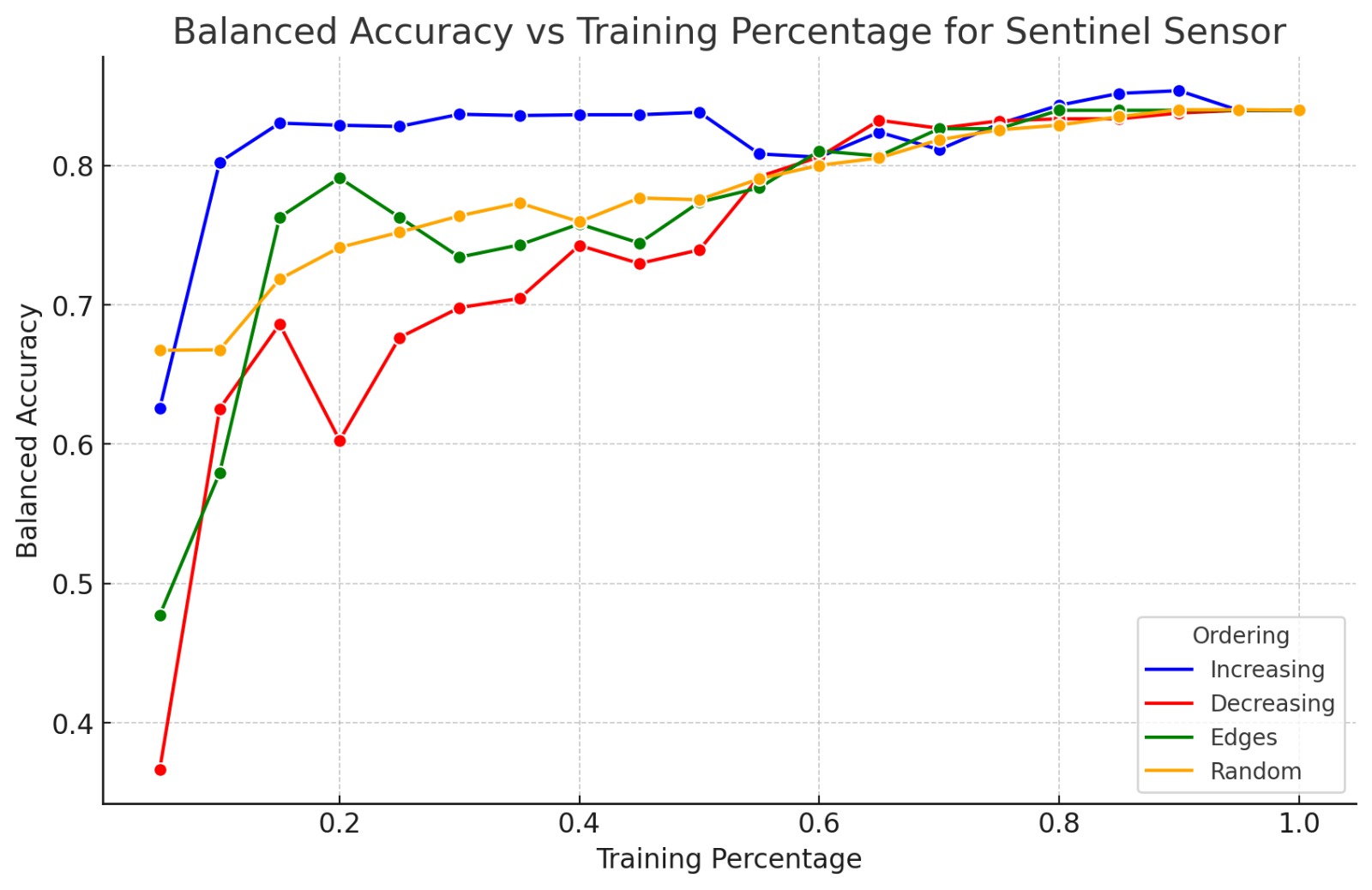}
\caption{Classification results of the sampling strategies for deforestation detection task.}
\label{fig:curves_sentinel}
\end{figure}

As presented in Section \ref{sec:FE}, three different sampling strategies were used to train the SVM technique using a linear kernel. In addition to these, a random sample selection strategy was also employed in this work, maintaining the same quantitative proportions of samples per class. Regarding the random selection strategy, to ensure a fair comparison, the model was performed by 10 times, and at the end, the arithmetic mean of the balanced accuracy was calculated. The results observed for the three different sampling strategies implemented in this study, along with the random approach, are presented in Figure~\ref{fig:curves_sentinel}.

When analyzing the model's behavior for each strategy with the respective training set, a notable advantage of the \textit{increasing} strategy is observed when using training sets with up to 55\% of the samples with the lowest entropy (highest confident samples). This characteristic indicates that through this strategy, it is possible to train models like SVM with very few samples from this domain, achieving significant balanced accuracy results. This fact can be especially useful when a large training set is not available. Conversely, the \textit{decreasing} strategy, where samples with the highest entropy are chosen first, did not prove as effective, only equating to the other strategies when using 60\% of the samples. The same could be applied to edge strategy, when the lowest and highest entropies were selected to organize the training set.

\section{Conclusion}
Citizen science campaigns are valuable tools that promote volunteer engagement in noble causes, the dissemination of globally relevant information, and the generation of data of considerable value. In this context, the research highlighted the ForestEyes project, in which non-specialized volunteers analyze and classify segments with potential deforestation in the Brazilian Legal Amazon. This classification process, at the end of each campaign, produces valuable labels for the segments, which can be used to train ML models such as SVM. As discussed in this paper, sample selection for effective training is a rather complex task. Considering that in the campaigns it is possible to measure the difficulty of classifying segments through the variability observed in their entropy, this research presented three different sample selection strategies for training the SVM based on this measure. Based on the results, it was observed that the approach based on selecting samples with lower entropy (\textit{increasing}) produces good results with few samples. Particularly, with only 10\% of the samples with the lowest entropy (9 from forest and 9 from non-forest), it was possible to achieve a balanced accuracy value that the other strategies only reached using around 70\% of the samples. For future work, it is intended to assess the robustness of this approach in other ML models, such as Multilayer Perceptron (MLP) neural networks and Random Forest (RF), exploring a greater parametric variation. Additionally, it is desired to compare the strategies tested in this work with other techniques from the literature, such as Margin Sampling.

\section*{Acknowledgements}
\textcolor{black}{The authors would like to thank UNIFESP and IFSULDEMINAS for institutional support;  CAPES and CNPq agencies for providing scholarship to H.R. and E.B.N.; and LNCC for providing high-performance computational resources. This publication uses data generated via the Zooniverse.org platform, development of which is funded by generous support, including a Global Impact Award from Google, and by a grant from the Alfred P. Sloan Foundation. This research is part of the INCT Future Internet for Smart Cities, funded by CNPq (\#465446/2014-0) and FAPESP (\#2014/50937-1, \#2015/24485-9,  \#2018/23908-1, \#2019/26702-8, \#2023/00811-0, and \#2023/00782-0).}

\vspace{-0.1cm}


\end{document}